\documentclass{article}

\newcommand{\mname}{Robustmix}
\usepackage{algorithm}
\usepackage{algorithmic}
\usepackage{amsmath}
\usepackage{amssymb}
\usepackage{mathtools}
\usepackage{amsthm}

\usepackage[textsize=tiny]{todonotes}
    \usepackage[final]{neurips_distshift_2022}

\usepackage[utf8]{inputenc} % allow utf-8 input
\usepackage[T1]{fontenc}    % use 8-bit T1 fonts
\usepackage{hyperref}       % hyperlinks
\usepackage{url}            % simple URL typesetting
\usepackage{booktabs}       % professional-quality tables
\usepackage{amsfonts}       % blackboard math symbols
\usepackage{nicefrac}       % compact symbols for 1/2, etc.
\usepackage{microtype}      % microtypography
\usepackage{xcolor}         % colors

\title{\mname{}: Improving Robustness by Regularizing the Frequency Bias of Deep Nets}

\author{%
  Jonas Ngnawé\thanks{These authors contributed equally, work done during AI residency at Google.} \\
  Université Laval and Mila-Quebec AI Institute\\
  \texttt{ngnawejonas@gmail.com} \\
   \And
  Marianne N. Abemgnigni\footnotemark[1] \\
  University of Göttingen\\
   \texttt{nmabemgnigni@aimsammi.org} \\
  \AND
    Jonathan Heek\\
  Google AI \\
   \texttt{jheek@google.com} \\
   \And
   Yann Dauphin \\
  Google AI \\
   \texttt{ynd@google.com} \\
}

\begin{document}

\maketitle

\begin{abstract}
    Deep networks have achieved impressive results on a range of well-curated benchmark datasets. Surprisingly, their performance remains sensitive to perturbations that have little effect on human performance. In this work, we propose a novel extension of Mixup called \mname{} that regularizes networks to classify based on lower-frequency spatial features. We show that this type of regularization improves robustness on a range of benchmarks, such as ImageNet-C and Stylized ImageNet. It adds little computational overhead and does not require a priori knowledge of a large set of image transformations. We find that this approach further complements recent advances in model architecture and data augmentation, attaining a state-of-the-art mean corruption error (mCE) of 44.8 with an EfficientNet-B8 model and RandAugment, which is a reduction of 16 mCE compared to the baseline.
\end{abstract}

\section{Introduction}

Deep neural networks have achieved state-of-the-art accuracy across a range of benchmark tasks such as image segmentation \citep{ren2015faster} and speech recognition \citep{hannun2014deep}. These successes have led to the widespread adoption of neural networks in many real-life applications. However, while these networks perform well on curated benchmark datasets, their performance can suffer greatly in the presence of small data corruptions \citep{szegedy2013intriguing,goodfellow2014explaining,moosavi2017universal,athalye2018synthesizing,hendrycks2019benchmarking}. This poses significant challenges to the application of deep networks.

\citet{hendrycks2019benchmarking} show that the accuracy of a standard model on ImageNet can drop from 76\% to 20\% when evaluated on images corrupted with small visual transformations. This shows modern networks are not robust to certain small shifts in data distribution. That is a concern because such shifts are common in many real-life applications. Secondly, \citet{szegedy2013intriguing} show the existence of \emph{adversarial} perturbations, which are imperceptible to humans but have a disproportionate effect on the predictions of a network. This raises significant concerns about the safety of using deep networks in critical applications such as self-driving cars \citep{sitawarin2018darts}.

These problems have led to numerous proposals to improve the robustness of deep networks. Some of these methods, such as those proposed by \citet{hendrycks2019augmix}, require a priori knowledge of the visual transformations in the test domain. Others, such as \citet{geirhos2018imagenet} use a deep network to generate transformations which comes with significant computation cost.

This paper proposes a new data augmentation technique to improve the robustness of deep networks by regularizing frequency bias. This new regularization technique is based on Mixup. It has many advantages compared to related robustness regularizers: (1) it does not require knowledge of a large set of priori transformations, (2) it is inexpensive, and (3) it doesn't have many hyper-parameters.  The key idea is to bias the network to rely more on lower spatial frequencies to make predictions. 

We demonstrate on ImageNet-C that this method works well with recent advances and reaches a state-of-the-art mCE of 44.8, 85.0 clean accuracy with EfficientNet-B8 and RandAugment\citep{cubuk2019randaugment}. This is an improvement of 16 mCE compared to the baseline EfficientNet-B8 and matches ViT-L/16 \citep{dosovitskiy2020image}, which is trained on $300\times$ more data. Our implementation of the method with DCT transform adds negligible overhead in our experiments. We find that \mname improves accuracy on Stylized-ImageNet by up to 15 points, and we show that it can increase adversarial robustness.

% \begin{figure}[ht]
% \centering
% \includegraphics[width=\columnwidth]{images/scatter_mce_vs_error_all.pdf}
% \caption{Clean error vs mCE for various models. Our EfficientNet model trained with \mname{} obtains the best combination of accuracy and mCE.}
% \label{fig:acc_vs_mce_cloud}
% \end{figure}

\section{Related Work}

The proposed approach can be seen as a generalization of Mixup \citep{zhang2017mixup}, a data augmentation method that regularizes models by training them on linear interpolations of two input examples and their respective labels. 
These new examples are generated as follows
\begin{align}
  \tilde{x} &= \texttt{mix}(x_1, x_2, \lambda),\qquad \text{where~} x_1, x_2 \text{~are~input images}\nonumber\\
  \tilde{y} &= \texttt{mix}(y_1, y_2, \lambda),\qquad \text{where~} y_1, y_2 \text{~are~labels}\nonumber
\end{align}

with $\texttt{mix}$ being the linear interpolation function
\begin{align}
    \texttt{mix}(x_1, x_2, \lambda) = \lambda x_1 + (1 - \lambda) x_2
\end{align} where $\lambda \sim \mathrm{Beta}(\alpha, \alpha)$, $\alpha$ is the Mixup coefficient hyper-parameter.

\citet{zhang2017mixup} show that Mixup improves the accuracy of networks and can also improve the robustness of the network.
In the past years, several versions of Mixup were proposed with application in Fairness \citep{chuang2021fair}, 3D reconstruction \citep{cheng2022pose}, semi-supervised learning \citep{NEURIPS2019_f708f064}, as well as robustness (\citep{mai2021metamixup, yun2019cutmix, faramarzi2020patchup, kim2020puzzle, pmlr-v97-verma19a}). The novel version we propose here is frequency-based and does not include additional learnable parameters.

Augmix \citep{hendrycks2019augmix} is a data augmentation technique to improve robustness by training on a mix of known image transformations. It adds little computational overhead but requires knowledge of a diverse set of domain-specific transformations. \cite{hendrycks2019augmix} mixes a set of 9 different augmentations to reach $68.4$ mCE on ImageNet. In contrast, the proposed method does not rely on specific image augmentations but on the more general principle that natural images are a kind of signal where most of the energy is concentrated in the lower frequencies.

The idea of frequency filtering is popular in Deep learning frameworks and has numerous applications, including unsupervised domain adaptation (\cite{yang2020fda}) and adversarial perturbation attacks (\cite{https://doi.org/10.48550/arxiv.1809.08758, li2021f}). Unlike the latter papers, which focus on measuring the accuracy of a model after an adversarial attack, we focus on common (noise) corruptions by measuring mCE as a robustness assessment.

\citet{zhang2019making} uses low pass filters directly inside the model to improve the frequency response of the network.\cite{https://doi.org/10.48550/arxiv.1903.06256} uses a differentiable neural network to extract textual information from images without modelling the lower frequency. Our method also uses low-pass filtering but does not entirely remove high-frequency features. Additionally, we only use frequency filtering during training; therefore, no computational overhead is incurred during evaluation.

\section{Method}

\begin{figure*}[ht]
\centering
\includegraphics[trim={0 4cm 0 3cm},clip,width=\textwidth]{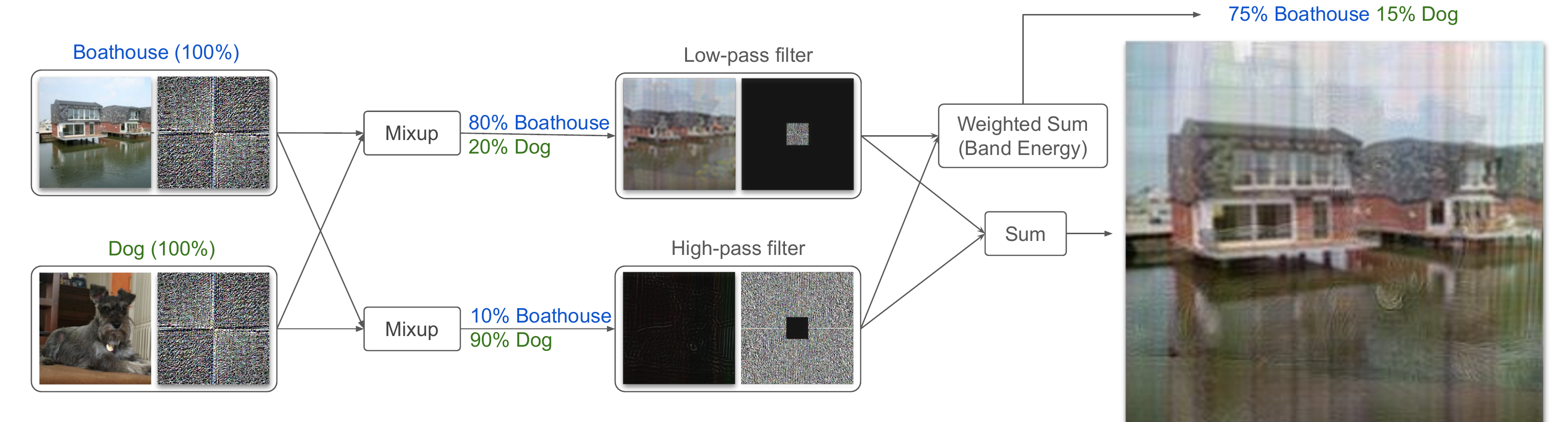}
 \caption{Illustration of the method. To better illustrate the method, we display the Fourier spectrum of the images next to them. We can see that even though 90\% of the higher frequencies belong to the image of a dog, \mname assigns more weight to the boathouse label because it assigns more weight to the lower frequencies.}
 % [\href{https://docs.google.com/presentation/d/1XLnBV3zKAn6FE_t4vDo2ZcsZbXm33mWLLscBUhqifVA/edit?usp=sharing&resourcekey=0-wRJITa9Gk78bcdbygQ1X5A}{slides}]
\label{fig:illus}
\end{figure*}

% \todo[inline]{Some readers may not have context about what high and low frequencies are or look like. It might be useful to have some pictures illustrating that. This can be referenced in the motivation section.}

In this section, we introduce a novel extension of Mixup called \emph{\mname{}} that increases robustness by regularizing the network to focus more on the low-frequency features in the signal.

{\bf Motivation} \citet{wang2020high} suggest that convolutional networks trade robustness for accuracy in their use of high-frequency image features. Such features can be perturbed in ways that change the model's prediction, even though humans cannot perceive the change. This can lead models to make puzzling mistakes, such as with adversarial examples. We aim to increase robustness while retaining accuracy by regularizing how the model uses high-frequency information.

% {\bf Proposal} Our proposal is to extend Mixup by mixing low and high-frequency bands separated by a cutoff frequency $c$. This allows use to regularize the model's bias to specific frequencies in the input image.
{\bf \mname{}} We propose to regularize the model's sensitivity to each frequency band by extending Mixup's linear interpolations with a new type of band interpolation. The key insight is that we can condition the sensitivity to each band using images that mix the frequency bands of two different images. Suppose that we mix the lower-frequency band of an image of a boathouse with the high-frequency band of an image of a dog. We can encourage sensitivity to the lower band by training the model to predict "dog" for this mixed image. However, this approach is too simplistic because it completely disregards the impact of the image in the high band. Indeed, the ablation study in section \ref{sec:ablation} shows that it is insufficient.

\begin{figure}
\centering
\vspace{-5mm}
\includegraphics[width=\columnwidth]{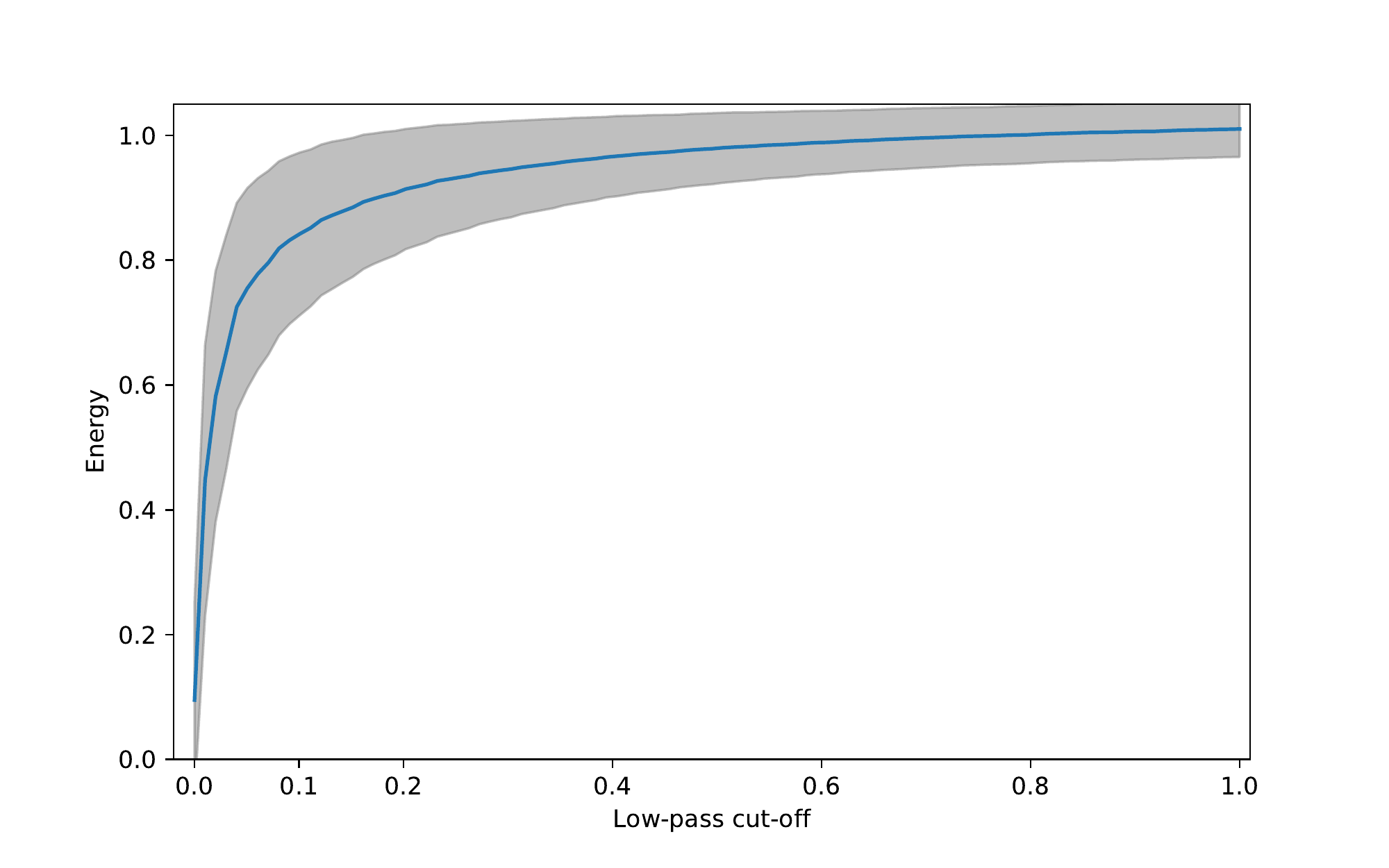}
\caption{Plot of the cumulative energy in ImageNet images as a function of the frequency cutoff.}
% [\href{https://colab.corp.google.com/drive/1a0g40R1Rxywfz2Nw8itpXEuYt0MC9jIW?usp=sharing}{colab}]
\vspace{-2mm}
\label{fig:energy}
\end{figure}
%\todo[inline]{Jonas: The axis are a bit weird. Make the figure start the x-axis and y-axis actually start at 0.}

Instead, we interpolate the label of such mixed images according to an estimate of the importance of each frequency band. We propose using the relative amount of energy in each band to estimate the importance. Thus the sensitivity of the model to high-frequency features will be proportional to their energy contribution in natural images. As shown in Figure \ref{fig:energy}, most of the spectral energy in natural images is concentrated in the lower end of the spectrum. This should limit the ability of high-frequency perturbations to change the prediction unilaterally. 

Furthermore, we use linear interpolations of images like in mixup within each band instead of raw images. This closely reflects the more common case where the features in the bands are merely corrupted instead of entirely swapped. It also has the benefit of encouraging linearity inside the same frequency band. % This linear and band interpolation gives Doublemix its name.

Specifically, the mixing formula for \mname{} is given by
\begin{align}
    \tilde{x} &= \texttt{Low}(\texttt{mix}(x_1, x_2, \lambda_L), c) + \texttt{High}(\texttt{mix}(x_1, x_2, \lambda_H), c) \label{eq:xrmixed}\\ %\nonumber\\
    \tilde{y} &= \lambda_c \texttt{mix}(y_1, y_2, \lambda_L) + (1 - \lambda_c)\texttt{mix}(y_1, y_2, \lambda_H) \label{eq:yrmixed}%\nonumber
\end{align}
where $\lambda_L,\lambda_H \sim \mathrm{Beta}(\alpha, \alpha)$, $\alpha$ is the Mixup coefficient hyper-parameter, and $\texttt{Low}(\cdot, c), \texttt{High}(\cdot, c)$ are a low pass and high pass filter respectively with a uniformly sampled cutoff frequency $c \in [0,1]$. And $\lambda_c$ is the coefficient that determines how much weight is given to the lower frequency band. It is given by the relative amount of energy in the lower frequency band for natural images
\begin{align}
    \lambda_c = \frac{E[\|\texttt{Low}(x_i, c)\|^2]}{E[\|x_i\|^2]}.
\end{align}
This coefficient can be efficiently computed on a mini-batch of examples.

% { \color{blue}
% In order to achieve this, we first compute the mixing in each band
% \begin{align}
%     (\tilde{x}_L, \tilde{y}_L) &= \texttt{mixup}(\texttt{L}(x_i, c), \texttt{L}(x_j, c), \lambda_L) \nonumber\\
%     (\tilde{x}_H, \tilde{y}_H) &= \texttt{mixup}(\texttt{H}(x_i, c), \texttt{H}(x_j, c), \lambda_H) \nonumber
% \end{align}
% where $\lambda_L,\lambda_H \sim \mathrm{Beta}(\alpha, \alpha)$, $\alpha$ is the Mixup coefficient hyper-parameter, and $\texttt{L}(\cdot, c), \texttt{H}(\cdot, c)$ are a low pass and high pass filter respectively with a uniformly sampled cutoff frequency $c \in [0,1]$. The  final augmented image is given by
% \todo[inline]{Explain that we sometimes have a lower bound for the cutoff frequency so not [0, 1] but [c, 1]}

% \begin{align}
%     \tilde{x} &= \tilde{x}_L + \tilde{x}_H\nonumber\\
%     \tilde{y} &= \lambda_y\tilde{y}_L + (1 - \lambda_y)\tilde{y}_H\nonumber
% \end{align}
% where $\lambda_y$ is the coefficient that determines how much weight is given to the lower frequency band. It is given by the relative amount of energy in the lower frequency band
% \begin{align}
%     \lambda_y = \frac{E[\|\texttt{L}(x_i, c)\|^2]}{E[\|x_i\|^2]}
% \end{align}
% }

{\bf Implementation} Computational overhead is an important consideration for data augmentation techniques since training deep networks is computationally intensive, and practitioners have limited computational budgets. We note that many popular techniques such as Mixup \citep{zhang2017mixup} add little overhead.

The frequency separation is implemented using a Discrete Cosine Transform (DCT) to avoid the complex multiplication required by a Discrete Fourier Transform. We directly multiply the images with the 224x224 DCT matrix because the spatial dimensions are relatively small, and (non-complex) matrix multiplication is well-optimized on modern accelerators. A batch of images is transformed into frequency space and the low and high-pass filtered images must be transformed back to image space. Additionally, we must apply the DCT transform over the x and y dimensions separately. Thus, 6 DCT matrix multiplications are required, resulting in $0.2$  GFLOPs per image. In contrast, just the forward pass of ResNet50 requires $3.87$ GFLOPs \citep{hasanpour2016lets}.

In our implementation of \mname{}, we reorder commutative operations (low pass and mixing) to compute the DCT only once per minibatch. The pseudocode is provided in Algorithm \ref{alg:robustmix}, where $\mathrm{reverse}$ is a function that reverses the rows of its input matrix.

\begin{algorithm}
\caption{\mname{}}\label{alg:robustmix}
\begin{algorithmic}
\STATE {\bfseries Input:} Minibatch of inputs $X\in\mathbb{R}^{N\times H\times W\times D}$ and labels $Y\in\mathbb{R}^{N\times C}$, $\alpha \in \mathbb{R}$
\STATE {\bfseries Output:} Augmented minibatch of inputs $\tilde{X}\in\mathbb{R}^{N\times W \times H\times D}$ and labels $\tilde{Y}\in\mathbb{R}^{N\times C}$
\STATE $\lambda_L,\lambda_H \sim \mathrm{Beta}(\alpha, \alpha)$ and $c \sim U(0, 1)$
\STATE $L \gets \texttt{Low}(X, c)$
\STATE $H \gets 1 - L$
\STATE $\lambda_c \gets \frac{\|L\|^2}{\|X\|^2}$
\STATE $\tilde{X} \gets \texttt{mix}(L, \mathrm{reverse}(L), \lambda_L) + \texttt{mix}(H, \mathrm{reverse}(H), \lambda_H)$
\STATE $\tilde{Y} \gets \texttt{mix}(Y, \mathrm{reverse}(Y), \lambda_c * \lambda_L + (1 - \lambda_c) * \lambda_H)$
\end{algorithmic}
\end{algorithm}

% At first glance, it may seem that the proposed method adds significant overhead since the low-pass filtering with DCT has $O(N\log N)$ complexity. However, this operation is embarrassingly parallel and only executed once per input batch. We find that \mname adds less than $1\%$ overhead to the computation of a training step for training EfficientNet-B5 on ImageNet.

\section{Results}

\subsection{Datasets and Metrics}

The results presented in this paper rely on the mCE measurement on ImageNet-C, the clean accuracies on ImageNet and Stylized-ImageNet (SIN), and the shape bias on SIN. These measurements are found in a range of papers studying robustness \citep{hendrycks2019benchmarking, hendrycks2019augmix, geirhos2018imagenet, laugros2020addressing}. 
% Our approach puts together the corresponding values reported in several papers or computed by ourselves for the sake of comparison of our method with other ones.
%the usual benchmarking dataset and metrics for the robustness literature
 The Stylized-ImageNet benchmark aims to distinguish between a bias towards shape or texture. We believe our results on Stylized-ImageNet complement the standard robustness results because they show that the inductive bias is more human-like in the sense that it is more sensitive to shape than texture.

{\bf ImageNet.}
ImageNet \citep{imagenet} is a classification dataset that contains 1.28 million training images and 50000 validation images with 1000 classes. We evaluate the common classification accuracy, which is referred to as clean accuracy. We use the standard Resnet preprocessing, resulting in images of size 224x224 \citep{resnet}.
%a scale augmentation that consists of resizing each image with its shorter size randomly sampled in [256, 480]; followed by a 224x224 crop from the resulting image or its horizontal flip, with per-pixel mean subtracted, and the standard color augmentation.
The standard models without any additional data augmentation process, will be qualified as the baseline.

%\todo[inline]{We don't need to explain this setup just mention it is the same setup as standard paper X}

{\bf ImageNet-C.}
This dataset comprises 15 types of corruption drawn from four main categories: noise, blur, weather, and digital \citep{hendrycks2019benchmarking}. These corruptions are applied to the validation images of ImageNet at five different intensities or levels of severity. Following \citep{hendrycks2019benchmarking}, we evaluate the robustness of our method by reporting its \textbf{mean corruption error (mCE)} normalized with respect to AlexNet errors: \[\text{mCE} = \frac{\sum\limits_{\text{corruption } c} {\text{CE}}_c}{\text{Total Number of Corruptions}}, \; \] \[\text{with } {\text{CE}}_c = \frac{\sum\limits_{\text{severity } s}E_{c,s}}{\sum_{s} E_{c,s}^{\mathrm{AlexNet}}}\]

{\bf Stylized-ImageNet.}
Stylized-ImageNet (SIN) is constructed from ImageNet by replacing the texture in the original image using style transfer, such that the texture gives a misleading cue about the image label \citep{geirhos2018imagenet}. The 1000 classes from ImageNet are reduced to 16 shape categories, for instance, all labels for dog species are grouped under one "dog" label, same for "chair", "car", etc. There are 1280 generated cue conflict images (80 per category). We evaluate the classification accuracy (SIN accuracy) and measure the model's shape bias with SIN.
Following \citet{geirhos2018imagenet}, the model's bias towards shape versus texture is measured as
\[ \text{shape bias} = \frac{\text{correct shapes}}{\text{correct shapes + correct textures}}. \]

% i.e the ratio of correctly classified shapes over the total number of correctly classified images with cue-conflict (shape or texture) i.e images where shape and texture do not coincide.

\subsection{Experimental Setup}

We evaluated residual nets (ResNet-50  and ResNet-152) and EfficientNets (EfficientNet-B0, EfficientNet-B1, EfficientNet-B5, and EfficientNet-B8). Experiments were run on 8x8 TPUv3 instances for the bigger EfficientNets (EfficientNet-B5 and EfficientNet-B8), and the other experiments were run on 4x4 TPUv3 slices. For the Resnet models, we use the same standard training setup outlined in \cite{goyal2017accurate}. However, we use cosine learning rate \cite{loshchilov2016sgdr} with a single cycle for Resnets trained for 600 epochs.

\subsection{Robustness Results}

\begin{figure}[ht]
\centering
\includegraphics[width=\columnwidth]{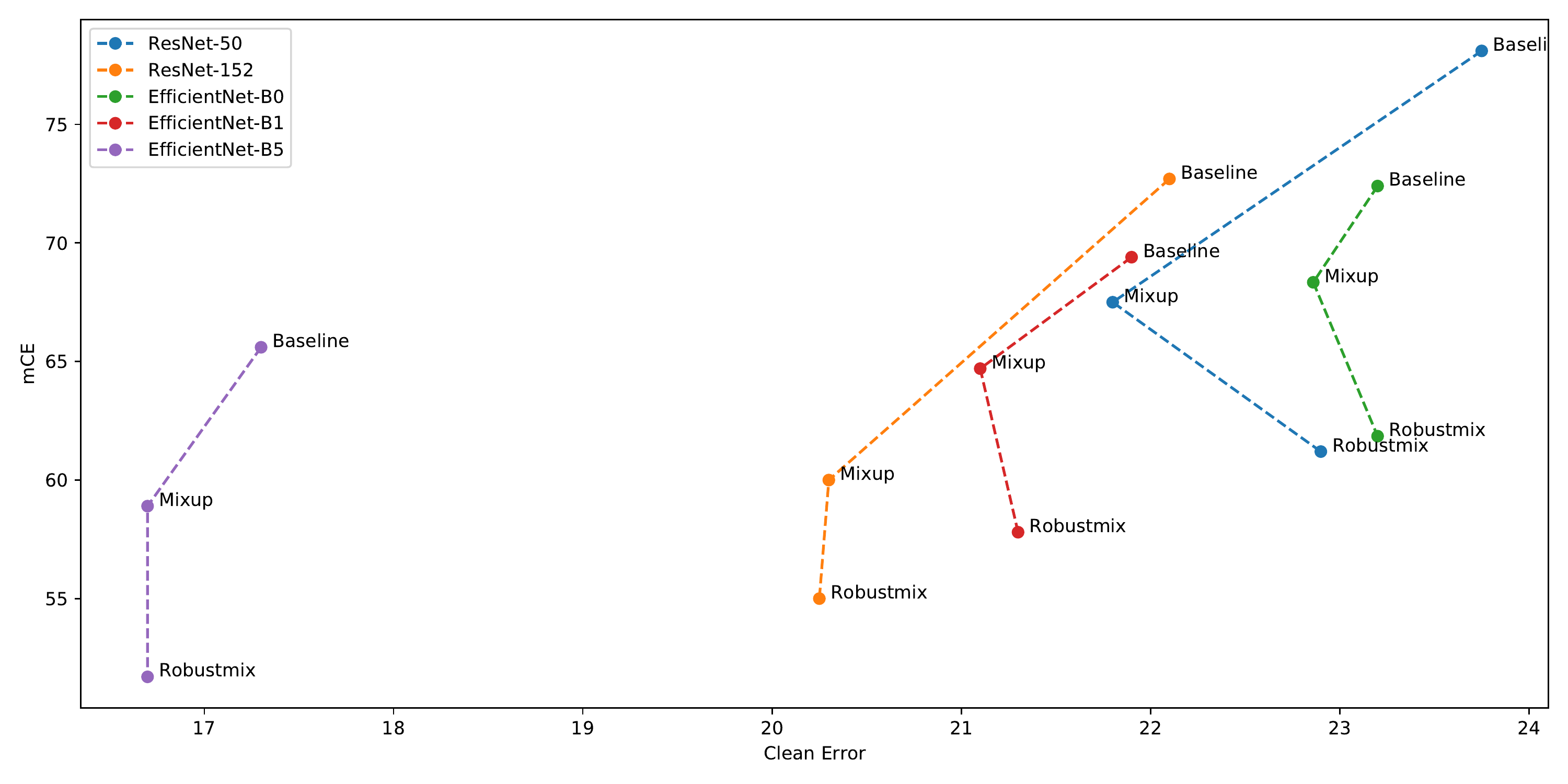}
\caption{Highlighting the tradeoff between mCE and Clean Error for various models.}
\label{fig:acc_vs_mce_tradeoff}
\end{figure}

{\bf ImageNet-C} First, we evaluate the effectiveness of the proposed method in improving the robustness to visual corruptions considered in ImageNet-C. In Table \ref{tresults}, we can see that \mname{} consistently improves robustness to the considered transformations, with a 15-point decrease in mCE over the baseline for ResNet-50. \mname{} with ResNet-50 achieves 61.2 mCE without degrading accuracy on the clean dataset compared to the baseline. In fact, we find a small improvement over the baseline of 0.8\% on the clean error. While Mixup yields a more significant gain of 1.9\% on the clean accuracy, we find that \mname{} improves mCE by up to 6 points more than Mixup. These results also compare favorably to Augmix, which needs to be combined with training on Stylized ImageNet (SIN) to reduce the mCE by 12 points. This improvement comes at a significant cost to accuracy due to the use of the Stylized ImageNet dataset. We also observe a similar trade-off between accuracy and robustness as we can observe in Figure \ref{fig:acc_vs_mce_tradeoff}. Mixup consistently produces lower clean error for smaller models, but the accuracy gap with \mname{} disappears as the model gets bigger.

% \todo[inline]{Break down the part explaining Fig 4 more}
While it is not directly comparable to ViT-L/16 due to its use of $300\times$ more data, EfficientNet-B8 with \mname and RandAugment has better robustness at $44.8$ mCE. It is also competitive with DeepAugment \citep{hendrycks2020many}, which requires training additional specialized image-to-image models on tasks such as super-resolution to produce augmented images. By comparison, our approach does not rely on extra data or extra-trained models.

\begin{table*}[ht]
\centering
\begin{tabular}{lllll}
\cline{1-5}
Method & \begin{tabular}[c]{@{}l@{}}Clean \\ Accuracy\end{tabular} & mCE & Size & \begin{tabular}[c]{@{}l@{}}Extra \\ Data\end{tabular} \\ \cline{1-5}
ResNet-50 Baseline (200 epochs) & 76.3 & 76.9 & 26M & 0 \\
ResNet-50 Baseline (600 epochs) & 76.3  & 78.1 & 26M & 0 \\
ResNet-50 BlurPool \citep{zhang2019making} & 77.0 & 73.4 & 26M & 0 \\
ResNet-50 Mixup    (200 epochs) &  77.5 & 68.1 & 26M & 0 \\
ResNet-50 Mixup    (600 epochs)& \textbf{78.2}   & 67.5  & 26M & 0 \\
ResNet-50 Augmix & 77.6   & 68.4 & 26M & 0 \\
ResNet-50 Augmix + SIN & 74.8   & 64.9 & 26M & 0 \\
% ResNet-50 \mname{} (200 epochs) & 77.0 & 63.8 & 26M & 0 \\
ResNet-50 \mname{} (600 epochs) & 77.1 & \textbf{61.2} & 26M & 0 \\
\hline
EfficientNet-B0 Baseline & 76.8 & 72.4  & 5.3M & 0 \\
EfficientNet-B0 Mixup ($\alpha = 0.2$) & \textbf{77.1}  & 68.3  & 5.3M & 0 \\
EfficientNet-B0 \mname{} ($\alpha=0.2$)  & 76.8 & \textbf{61.9} & 5.3M & 0 \\
\hline
EfficientNet-B1 Baseline & 78.1 & 69.4 & 7.8M & 0 \\
EfficientNet-B1 Mixup ($\alpha = 0.2$)  & \textbf{78.9} & 64.7 & 7.8M & 0 \\
EfficientNet-B1 \mname{} ($\alpha=0.2$)  & 78.7 & \textbf{57.8} & 7.8M & 0 \\
\hline
EfficientNet-B5 Baseline & 82.7 & 65.6 & 30M & 0 \\
EfficientNet-B5 Mixup ($\alpha = 0.2$)  & 83.3 & 58.9 & 30M & 0 \\
EfficientNet-B5 \mname{} ($\alpha=0.2$)  & 83.3 & 51.7 & 30M & 0 \\
EfficientNet-B5 RandAug+\mname{} ($\alpha=0.2$) & \textbf{83.8} & \textbf{48.7} & 30M & 0 \\ 
\hline
BiT m-r101x3 \citep{kolesnikov2020big} & 84.7 & 58.27  & 387.9M & 12.7M \\ 
\begin{tabular}[c]{@{}l@{}}ResNeXt-101 $32\times8d$+DeepAugment+AugMix  \\ \citep{hendrycks2020many}\end{tabular}& 79.9 & \textbf{44.5} & 88.8M & \begin{tabular}[c]{@{}l@{}}Extra  \\ models\end{tabular} \\ 
ViT-L/16 \citep{dosovitskiy2020image} & \textbf{85.2} & 45.5 & 304.7M & 300M \\
RVT-$B^{*}$ \citep{mao2022towards} & \textbf{82.7} & 46.8 & 91.8M & PAAS+Patch-wise\footnote{PAAS: Position-Aware Attention Scaling + a Simple and general patch-wise augmentation method for patch sequences.} 
\\
% RVT-$B^{*}$ \citep{mao2022towards} & \textbf{82.7} & 46.8 & 91.8M & PAAS
% Noisy Student & -- & 28.3 & - & 300M (from JFT)\\
\hline
EfficientNet-B8 Baseline & 83.4  & 60.8 & 87.4M & 0 \\
EfficientNet-B8 \mname{} ($\alpha=0.4$) & 84.4  & 49.8 & 87.4M & 0 \\
EfficientNet-B8 RandAug+\mname{} ($\alpha=0.4$)  & \textbf{85.0}  & \textbf{44.8} & 87.4M & 0 \\
\hline
\end{tabular}
\caption{Comparison of various models based on ImageNet accuracy and ImageNet-C robustness (mCE). The robustness results for BiT and ViT are as reported by \citet{paul2021vision}(Table 3).}
%DeepAugment and BiT are pretrained on ImageNet-21k which has 14M labeled training data (vs. 1.28M for ImageNet-1k). ViT extra data are from JFT-300M.
\label{tresults}
\end{table*}

Our experiments also show that Doublemix combines well with RandAugment (RA), further improving accuracy and mCE. We removed augmentations from RA that overlap with corruptions in ImageNet-C (contrast, color, brightness, sharpness, and Cut-out) \citep{hendrycks2019augmix}.

% With EfficientNet-B8 (88M parameters) and RA+Doublemix is comparable to ViT L-16 (304M parameters) having an accuracy of $85.05$ vs $85.15$ and mCE of $45.9$ vs $45.45$, and better than BiT m-r101x3 (387M parameters) having an accuracy of $84.7$ and $58.27$ of mCE as reported by \citet{paul2021vision}. Note that both ViT L-16 and BiT m-r101x3, in addition to being significantly larger in size, are trained with additional data.

In our cross-validation of  $\alpha$, we found small values less than $0.2$ perform poorly both on accuracy and mCE. Values of $\alpha$ such that $0.2 \leq \alpha \leq 0.5$ give not only the best accuracies and mCEs but also the best trade-off of mCE versus accuracy as bigger values of $\alpha$ have shown giving good values for accuracy but do not do as well on mCE.
In our experiments, we typically achieve good results with a frequency cutoff $c$ sampled between $[0, 1]$ as described in Algorithm \ref{alg:robustmix}. However, for ResNet-50 trained with a training budget that is too limited (200 instead of 600 epochs) and its smaller versions (ResNet-18 and ResNet-34), it can be beneficial to fix a minimum $c\geq\tau$ for the cutoff by sampling in the interval $[\tau, 1]$. The minimum cutoff determines the range at which band mixing will occur. We can remove band interpolation entirely and recover standard Mixup by setting $\tau=1$. For Resnet-50 with too few training epochs, we found that a good value for the minimum is $0.1$, but we found much better results can be achieved with 600 epochs without any modifications to Algorithm \ref{alg:robustmix}.

{\bf Stylized-ImageNet.}
% Figure \ref{fig:acc_vs_lhf} shows that our model is more robust to low-pass filters than the vanilla Mixup and the baseline. Since low frequencies in an image carry more of the visual features i.e what a human relies more on, and as such, they correspond to what determines the shape of an object in an image. Therefore, models trained with our method learn more shapes over texture unlike a deep neural network would naturally be tempted to do. 
We confirm that our method increases both accuracy on Stylized ImageNet and the shape bias as shown in table \ref{tab:sin_results}. For ResNet-50, \mname{} almost doubles the shape bias from baseline (from 19 to 37) and improves it by 63\% over Mixup, while relative improvements on SIN accuracy are 72\% and 33\% respectively over baseline and Mixup. The same observation is for EfficientNet-B5, which improves shape bias by nearly 50\% and SIN accuracy by almost 60\% over the baseline.
% and relative to mixup the improvements are 37\%  and 40\% respectively.

\begin{table*}[ht]
\resizebox{\columnwidth}{!}{%
\begin{tabular}{l|l|l|ll}
\hline
Method &
  Mixed Image &
  Label &
  \begin{tabular}[c]{@{}l@{}}Test\\ Accuracy\end{tabular} &
  mCE \\ \hline
\begin{tabular}[c]{@{}l@{}}\mname - Full\\ (inband mixups and energy weighting)\end{tabular} &
  Equation \ref{eq:xrmixed} &
  Equation \ref{eq:yrmixed} &
  77.1 &
  61.2 \\ \hline
\mname{} without energy weighting &
  Equation \ref{eq:xrmixed} &
  \begin{tabular}[c]{@{}l@{}}Equation \ref{eq:yrmixed}\\ with $\lambda_c$\\  replaced by $c$\end{tabular} &
  77.6 &
  67.7 \\ \hline
\begin{tabular}[c]{@{}l@{}}\mname{} without inband mixups\\ and with energy weighting\\ ($\lambda_L=1, \lambda_H=0$)\end{tabular} &
  $\texttt{Low}(x_1, c)  + \texttt{High}(x_2, c)$ &
  $\lambda_c y_1  + (1 - \lambda_c)y_2$ &
  68.6 &
  75.3 \\ \hline
\begin{tabular}[c]{@{}l@{}}\mname{} without inband mixups\\ and without energy weighing\\  ($\lambda_L=1, \lambda_H=0$\\  and cutoff c as label coefficient)\end{tabular} &
  $\texttt{Low}(x_1, c) + \texttt{High}(x_2, c)$ &
  $ c y_1 + (1 - c)y_2$ &
  74.8 &
  77.5 \\ \hline
\end{tabular}%
}
\caption{Comparison of \mname{} with simplified cases. The results are reported on ResNet50.}
\label{tab:ablation}
\end{table*}

\begin{table}[h]
\centering
\begin{tabular}{lll}
\hline
Method/Parameters & \begin{tabular}[c]{@{}l@{}}SIN \\ Accuracy\end{tabular} & \begin{tabular}[c]{@{}l@{}}Shape \\ Bias\end{tabular} \\ \hline
ResNet-50 Baseline & 15.6  & 19.25 \\ 
ResNet-50 Mixup & 20.1 & 22.7 \\ 
ResNet-50 \mname{}                       & \textbf{26.8} & \textbf{37.0} \\ \hline
EfficientNet-B5 Baseline & 25.3  & 44.4\\ 
EfficientNet-B5 Mixup & 28.75 &  48.3 \\ 
EfficientNet-B5 \mname{} & \textbf{40.3} & \textbf{66.1} \\ \hline
\end{tabular}
\caption{Accuracy and shape bias computed on Stylized ImageNet.}
% [\href{https://colab.corp.google.com/drive/1mVfdWC1qBsHhO0Bd6vlTTZ-Zvtqs7B9E?usp=sharing}{colab}]
\label{tab:sin_results}
\end{table}
%\todo[inline]{Please use either error or accuracy consistently (Table 2)}

\subsection{Ablation Study}\label{sec:ablation}
In order to measure the effect of \mname{}, we apply some simplifications to the image mixing and the labeling. The results are compiled in table \ref{tab:ablation}. It can be noticed from the first two lines that ablating the energy weighting results in a significant drop in mCE, even though there is a slight accuracy improvement. However, keeping the energy weighting but not applying the inband mixups is largely detrimental to accuracy and robustness. These results show that \mname achieves a better combination of mCE and accuracy than these ablations.

\subsection{Analysis and Discussion}
\begin{figure}[H]
\centering
\includegraphics[width=\columnwidth]{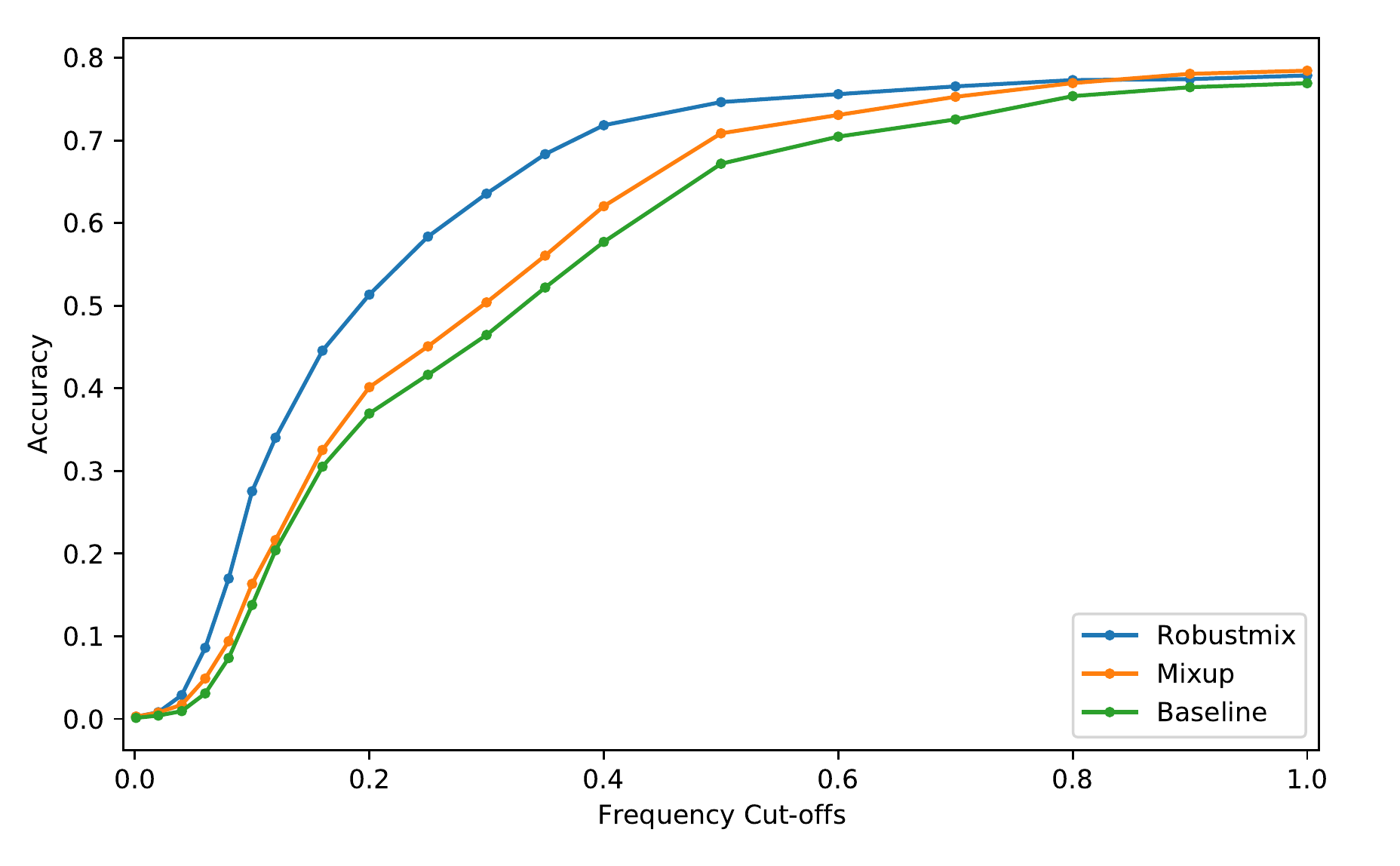}
\caption{Test accuracy on ImageNet samples passed through a low-pass filter with increasing cut-off. As expected, we observe that \mname is more robust to the removal of high frequencies than Mixup. The comparison is done here on ResNet-50 models.}

\label{fig:acc_vs_lhf}
\end{figure}

{\bf Low-frequency bias} In order to quantify the degree to which models rely on lower frequencies, we measure how much accuracy drops as we remove higher-frequency information with a low-pass filter.
Figure \ref{fig:acc_vs_lhf} shows that \mname is comparatively more robust to the removal of high frequencies. This indicates that models trained with \mname{} rely significantly less on these high-frequency features to make accurate predictions.

\section{Conclusion}

In this paper, we have introduced a new method to improve robustness called \mname{}, which regularizes models to focus more on lower spatial frequencies to make predictions. We have shown that this method yields improved robustness on a range of benchmarks, including ImageNet-C and Stylized ImageNet. In particular, this approach attains an mCE of 44.8 on ImageNet-C with EfficientNet-B8, which is competitive with models trained on $300\times$ more data. 

Our method offers a promising new research direction for robustness with several open challenges. We have used a standard DCT-based low-pass filter on images and an L2 energy metric to determine the contribution of each label. This leaves many alternatives to be explored, such as different data modalities like audio, more advanced frequency separation techniques like Wavelets, and alternative contribution metrics for mixing labels.

\section*{References}
\bibliographystyle{icml2021}
\bibliography{main}

% \appendix

% \section{Appendix}

\end{document}